\documentclass{article}
\usepackage{graphicx}
\usepackage[dvipsnames]{xcolor}
\usepackage{geometry}
\usepackage{hyperref}
\hypersetup{
  colorlinks   = true, 
  urlcolor     = blue, 
  linkcolor    = blue, 
  citecolor   = black 
}
\usepackage[utf8]{inputenc}
\usepackage[
backend=biber,
style=authoryear
]{biblatex}
\usepackage{cleveref}
\usepackage{siunitx}
\title{Geotechnical Parrot Tales (GPT): Harnessing Large Language Models in geotechnical engineering}

\author{Krishna Kumar}
\date{}

\begin{document}

\maketitle

\section{Introduction}
The rise of large language models (LLMs) has generated widespread interest due to their ability to answer questions, generate text/code, and summarize. One of the most popular LLM is OpenAI's ChatGPT, released in November 2022. ChatGPT is a probabilistic LLM built on OpenAI's GPT-3.5 (Generative Pretrained Transformers) with a chat user interface to the GPT model. Despite their impressive capabilities, ChatGPT can produce plausible-sounding but false outputs, blurring the line between facts and hallucinations~\parencite{bender2021dangers}.

Prompt engineering is creating effective inputs to help LLMs generate the desired output. Prompt engineering is crucial to mitigate these risks and harness the full potential of ChatGPT in geotechnical engineering. Prompt engineering refers to carefully designing input prompts to elicit accurate and valuable responses from LLMs~\parencite{petroni2020context}. By developing a deep understanding of ChatGPT's strengths and limitations, engineers can create prompts that effectively guide the model toward generating valuable insights while minimizing the risks of falsehoods and hallucinations.

In this article, we will explore ChatGPT and its applications in geotechnical engineering. We will discuss the challenges and pitfalls associated with these models and highlight the importance of prompt engineering in navigating these challenges to ensure reliable and accurate outcomes.

\section{Demystifying ChatGPT: The technical bits}
The groundbreaking paper ``Attention is All You Need" introduced Transformers~\parencite{vaswani2017attention}, the `T` in the GPT models. Unlike previous natural language models, transformers could look at the whole context. Imagine we are reading a book. As you progress through the sentences, you naturally anticipate the next word based on your understanding of language and the context provided by the story. Think of the Transformer as a dedicated reader who has read thousands of books and learned to recognize patterns in how words and sentences are structured. When encountering a new sentence or paragraph, it uses this knowledge to make educated guesses about what word should come next. For example, if the Transformer reads a sentence like "During site investigation, the engineer discovered that the soil has a high degree of...", it would likely predict the next word as something like "saturation" or "compaction," based on its understanding of common phrases and concepts in the geotechnical field. The Transformer model achieves this by utilizing layers of attention mechanisms. These attention mechanisms help the model focus on the most relevant words or phrases in the given context, just like how we might focus on specific details when trying to predict while reading a book.

\begin{figure}[ht]
    \centering
    \includegraphics[width = 0.8\linewidth]{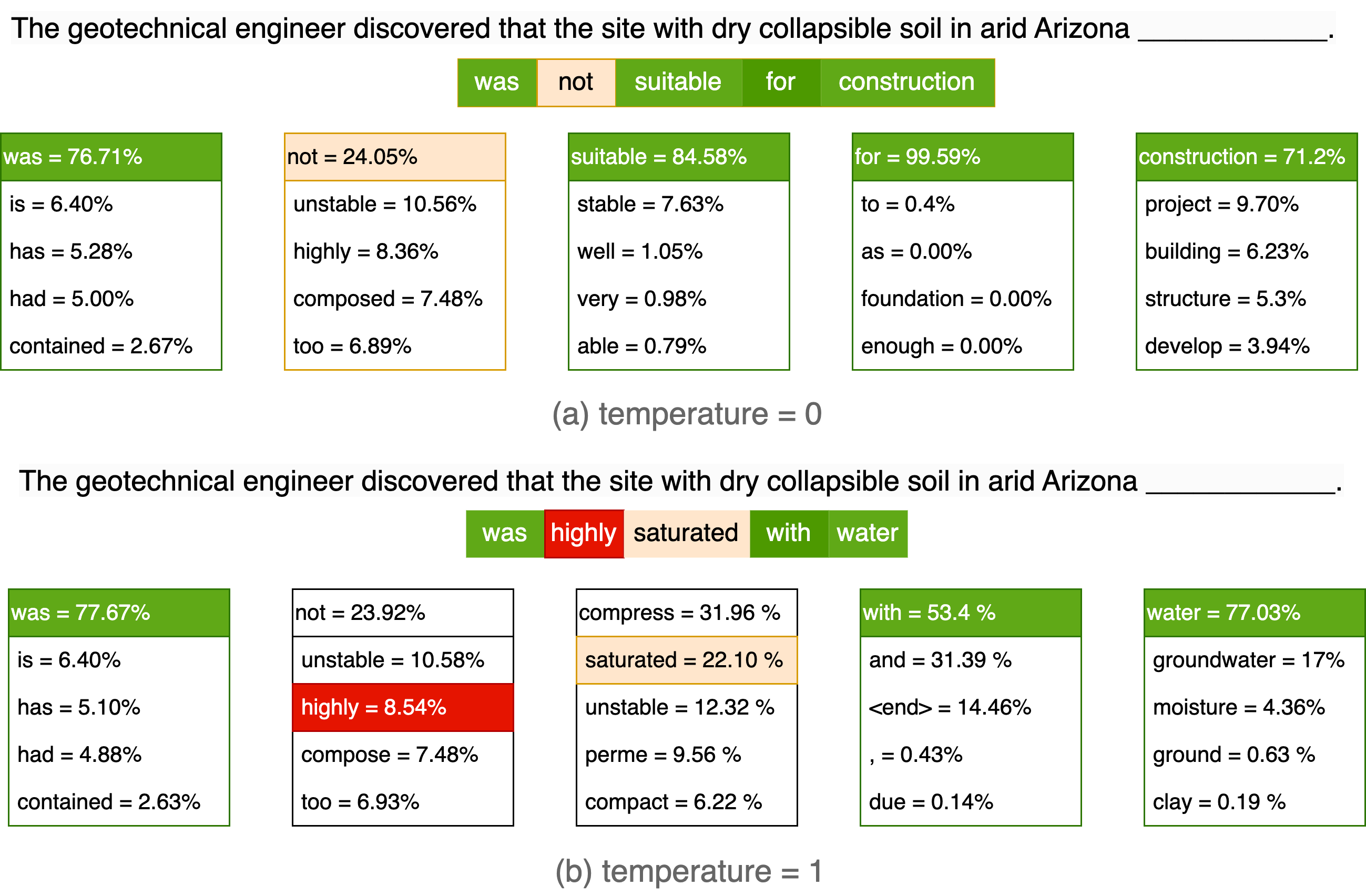}
    \caption{GPT text generation using transformers.}
    \label{fig:Transformers}
\end{figure}

\Cref{fig:Transformers}a illustrates how ChatGPT might complete the sentence, "The geotechnical engineer discovered that a site with dry collapsible soil in arid Arizona \underline{\hspace{1cm}}." GPT chooses each subsequent word based on probabilities derived from its extensive training corpus and the text generated thus far. This example selects the highest-ranking word at every step, resulting in the response, "... was not suitable for the proposed construction." While this method works well for short responses, it can take longer responses to appear robotic, stiff, and unimaginative. To address this, GPT provides various controls for adjusting word selection. One such control is the temperature setting. With a temperature of 0, GPT always chooses the word with the highest probability (\cref{fig:Transformers}a). However, by increasing the temperature to 1.0 (ChatGPT uses 0.7), GPT occasionally selects lower-ranked words, producing text that often feels more natural and inventive.~\Cref{fig:Transformers}b demonstrates this: a non-zero temperature setting generates an impossible and inaccurate response "was highly saturated with water" from the same input prompt. A zero temperature means the GPT model will be more consistent (the output will not change regardless of how often you run the same query), but it may still be wrong. Most scientific applications would require a temperature of zero to guarantee consistent results. 

ChatGPT is susceptible to hallucinations, meaning that it can occasionally generate responses that are not based on factual information or are unrelated to the context. These hallucinations can happen even when the temperature parameter is zero. A model's capability refers to its objective or goal, while alignment focuses on the desired behavior of the model compared to its actual training. ChatGPT's goal is to produce human-like text based on input, while its alignment aims to ensure that generated responses meet the expectations and values of its users. Models such as GPT-3 are capable but misaligned, as their training does not always capture higher-level meaning. This creates a noticeable gap between how these models are trained and how we want to use them. ChatGPT employs Reinforcement Learning from Human Feedback to better align with human values and expectations. GPT-3 has 175 billion parameters, while GPT-4 \parencite{bubeck2023sparks} is speculated to have 1 trillion parameters to encode all training information. ChatGPT generates plausible-sounding responses using these parameters alone, which leads to hallucinations, limited interpretability, and potentially biased output. Involving humans in the loop can introduce new biases. They may lack the required expertise for specific tasks (ChatGPT trainers may not include experts in Geotechnical Engineering), resulting in less accurate feedback.

ChatGPT can be described as a stochastic parrot because it learns from vast amounts of text data and generates human-like text based on the patterns it identifies. However, due to its training strategies, it may produce inconsistent, biased output or lack deeper understanding, much like a parrot that can mimic human speech but does not fully grasp the meaning behind the words. While ChatGPT is competent, addressing hallucination and alignment issues remains crucial to ensure its outputs align with human values and expectations.

\section{Applications of ChatGPT to Geotechnical Engineering}
Even with the current advancements in GPT-4, the models will hallucinate, i.e., lie or confidently make things up. Although ChatGPT is widely used to showcase its generative power, like writing emails and reports, we have not explored the innovative use cases of LLMs in engineering.  The best use of ChatGPT is not as an intelligent answering machine but as a reasoning engine or a ``calculator of words.” We will explore advanced ChatGPT use cases that can accelerate geotechnical engineering design and research.

\subsection{Prompt Engineering}
ChatGPT has a huge potential to accelerate the adoption of data-driven methods and computational modeling in Civil and Geotechnical Engineering. When we provide ChatGPT with a specific request called a ``prompt," it uses its vast knowledge to create a suitable response. However, crafting an effective prompt is crucial, as it guides the GPT in understanding what we want – ``prompt engineering." A well-designed prompt ensures GPT comprehends the task and delivers accurate, efficient, and functional code. Context is vital in prompt engineering for code generation because it helps the ChatGPT to better understand the specific requirements and nuances of the task at hand. Providing sufficient context allows the model to generate accurate and relevant code for the problem we are trying to solve. Without adequate context, ChatGPT might generate code that appears functional but lacks the necessary details to address the issue effectively. 

\begin{figure}
    \centering
    \includegraphics[width=\linewidth]{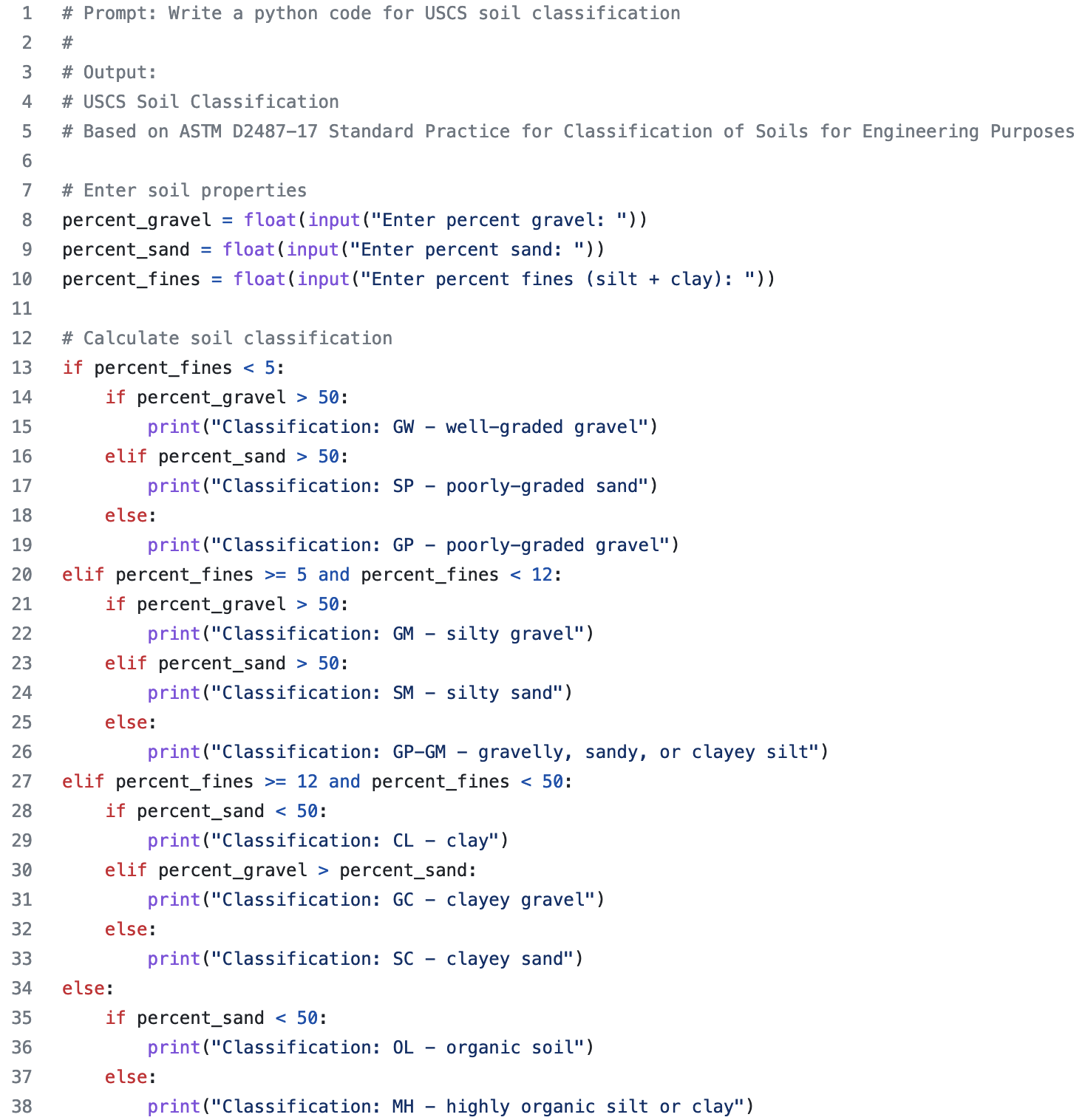}
    \caption{ChatGPT response to write a Python code for soil classification.}
    \label{fig:Soilclass}
\end{figure}

Consider the prompt:\textit{ \color{RedOrange}`Write a Python code for USCS soil classification.'}  In this case, ChatGPT hallucinates and responds with an incorrect code without including Atterberg’s limit, yet indicating it wrote the code based on ASTM D2487-17 (see~\cref{fig:Soilclass}). The prompt assumes that ChatGPT possesses all the necessary knowledge to accomplish the task. However, users should primarily use ChatGPT as a translator, providing the necessary context and knowledge to accomplish the task. We can provide a context by including ``\textit{Context: }” in the chat prompt and adding the relevant information. For the USCS soil classification, we will treat ChatGPT as a translator from English to Python. 

\noindent
\color{RedOrange}{\\\textbf{Prompt:} Write a Python code for USCS soil classification using the following conditions and using the A-line equation as 0.73 * (LL - 20).
\\\textbf{Context: }
\\\textbf{Inputs: }\% passing sieve 4 and 200, D10, D30, D60, Liquid limit (LL), and Plastic Limit (PL).
\\Compute auxiliary variables: Plasticity Index $(PI) = LL - PL$, $Cu = D60/D10$, and $Cc = D30^2/(D10 * D60)$
\\$<$Showing only Gravel classification for brevity$>$

\begin{enumerate}
    \item If the percentage passing through the No. 200 sieve (Sieve200) is less than or equal to 50, the soil is coarse-grained: 
        \begin{enumerate}
        \item If the percentage of gravel is greater than the percentage of sand, classify the soil as follows: 
            \begin{enumerate}
            \item If the percentage passing through the No. 200 sieve is less than 5 and the coefficient of uniformity (Cu) is greater than 4, and the coefficient of curvature (Cc) is strictly between 1 and 3, classify the soil as "GW". 
            \item If the above conditions are not met, classify the soil as "GP".
            \end{enumerate}
        \item If the percentage passing through the No. 200 sieve is greater than 12, classify the soil as follows: 
            \begin{enumerate}
            \item If the plasticity index (PI) is below the A-line, classify the soil as "GM".
            \item If the plasticity index is above the A-line, classify the soil as "GC". 
            \item If the plasticity index is on the A-line, classify the soil as "GM-GC" (borderline case).
            \end{enumerate}
        \item For borderline cases $(5\% \leq \textit{}Sieve200 \leq12\%)$, consider the following classifications:
            \begin{enumerate}
            \item If Cu $>4$ and Cc is strictly between 1 and 3, check the plasticity index:  If PI is below the A-line, classify the soil as "GW-GM".  If PI is above the A-line, classify the soil as "GW-GC". If PI is on the A-line, classify the soil as "GW-GM-GC" (borderline case).
            \item If the conditions above are unmet, check the plasticity index: If PI is below the A-line, classify the soil as "GP-GM". If PI is above the A-line, classify the soil as "GP-GC". If PI is on the A-line, classify the soil as "GP-GM-GC" (borderline case).
            \end{enumerate}
        \end{enumerate}
\end{enumerate}
}

\color{black}{
Similarly, ChatGPT can summarize articles and reports, but it is important to provide a context. For example, a summarization prompt without a context, such as, \textcolor{RedOrange}{\textit{Write a summary of  ``Crocker, Kumar, and Cox. (2023) Using explainability to design physics-aware CNNs for solving subsurface inverse problems.'' }}only gives a hallucinated summary. ChatGPT does not know the paper, as its knowledge is limited to articles published before September 2021. Even if we ask ChatGPT to summarize a paper published before September 2021, without context, ChatGPT will hallucinate. Accurate summarizing is only possible when the whole paper is provided as context. To summarize a soil report, we could ask ChatGPT specific information that needs summarization \textcolor{RedOrange}{\textit{``Summarize the strength parameters of the clay layer in the soil report: $<$insert report$>$''}}.}

\subsection{A context-specific search engine}
This section offers an approach to proving the correct context to LLMs. In the following sections, we will use OpenAI’s GPT-3.5-turbo Application Programming Interface (API) rather than the chat interface for context-specific Q\&As. LLMs are reasoning engines that analyze textual information to extract answers based on context. Unless we provide an appropriate context, GPT will suffer from hallucinations. For example, if we ask GPT, \textit{\textcolor{red}{``What is the XML tag to store plastic limit in DIGGS?”}} we get the following incorrect response: \textcolor{ForestGreen}{``\textless TestType \textgreater Plastic Limit\textless /TestType\textgreater."} However, the XML tag test type is a generic tag GPT invented to store Plastic Limit, although GPT has been trained on this data and is aware of DIGGS.  For GPT to generate answers, we need to provide appropriate context. In this case, we need to identify the appropriate sections of the DIGGS schema and provide it as a context. The number of tokens (words) we can provide as context is limited (32k), and the compute cost is related to the number of tokens. Hence, it is important to find the appropriate context rather than providing the entire knowledge base, which may not be possible in many cases. How can we build a context-specific search engine that uses local knowledge bases to find the correct context for GPT to answer the question? 

\begin{figure}[ht]
    \centering
    \includegraphics[width = 0.8\linewidth]{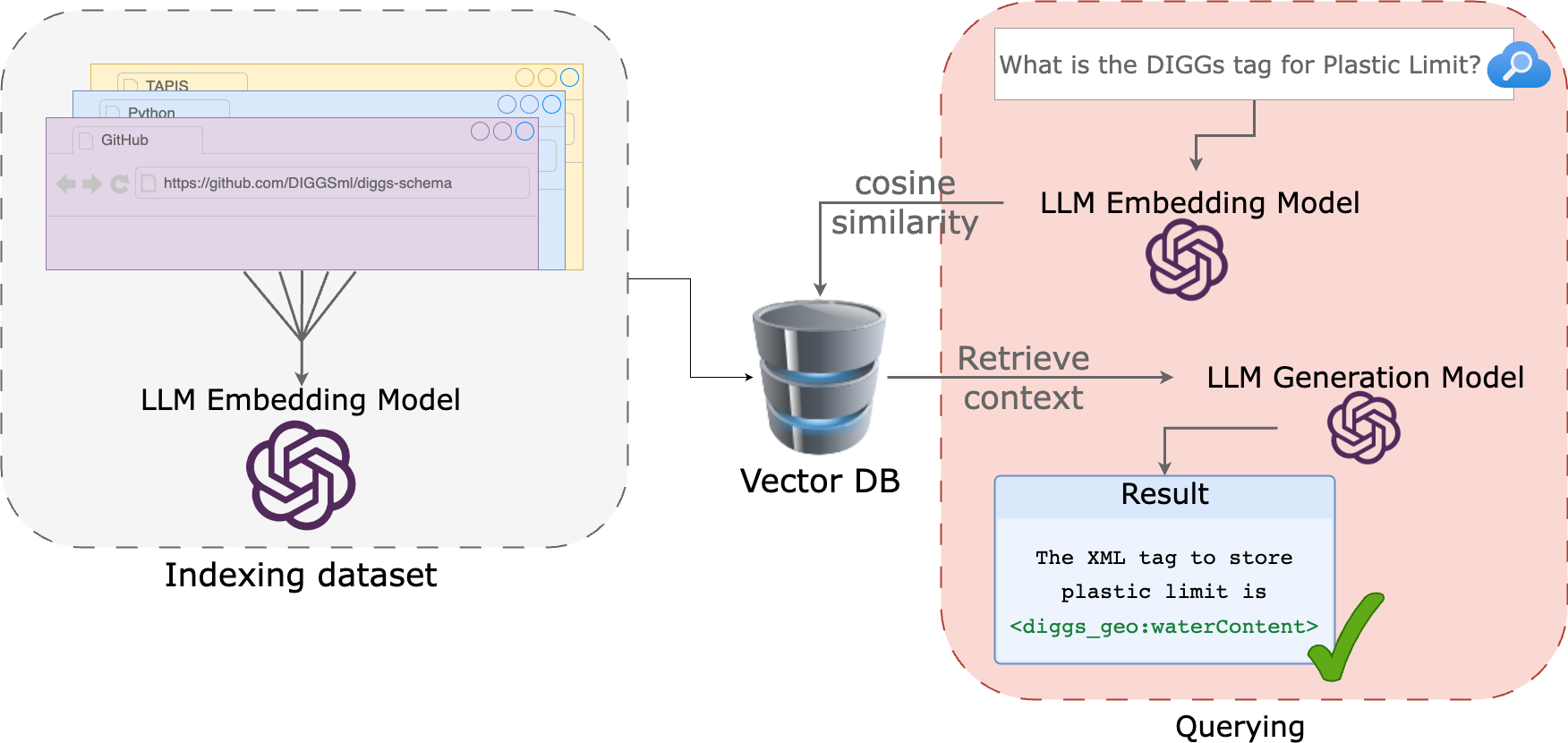}
    \caption{Context-specific semantic search engine using vector database and latent space embedding.
    \label{fig:semantic-search}
}
\end{figure}
\begin{figure}[ht]
    \begin{minipage}{0.48\textwidth}
    \textcolor{RedOrange}{Prompt: I have a list of three plastic limit trials: [11.9 \%, 11.7 \%, and 11.4\%]. Autoincrement trial number starting with one. Add a `gml:id` string for each trial. Create the XML file. }
    \\\\
    \textcolor{blue}{Response: Each plastic limit trial is represented by a `\textless diggs\_geo:PlasticLimitTrial\textgreater' element inside a `\textless diggs\_geo:plasticLimitTrial\textgreater' element. The `gml:id' attribute is set to ``tr1", ``tr2", and ``tr3" for the three trials, respectively. The `trialNo' element is auto-incremented starting from 1 for each trial, and the `waterContent' element contains the percentage values for the trials. The `isManual' element is set to true for all trials, but you can change it to false if needed.}
    \end{minipage}%
    \hspace{0.5cm}%
    \begin{minipage}[r]{0.45\textwidth}
    \includegraphics[width = \linewidth]{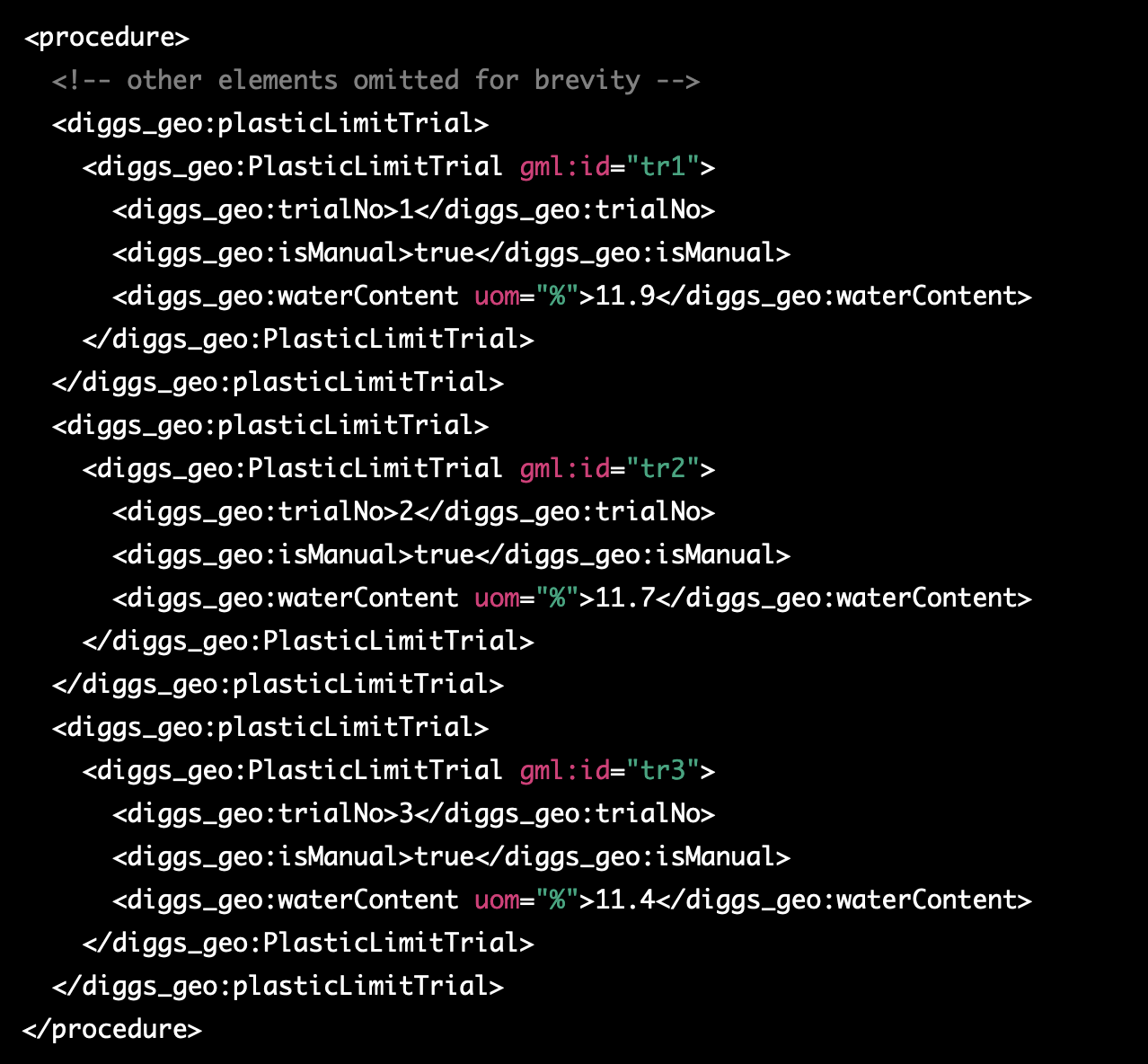}
    \end{minipage}
    \caption{Using GPT to generate DIGGS Schema.}
    \label{fig:diggs}
\end{figure}

Semantic search aims to understand the intent and contextual meaning of search queries, providing more accurate and relevant results. Figure 3 shows an overview of the semantic search engine architecture. Using OpenAI's text embedding model, we convert the textual knowledge base (e.g., DIGGS schema) into context vectors (see~\cref{fig:semantic-search}). The context vectors of the knowledge base are stored in a fast-accessible vector database like FAISS. When a user enters a search term, we convert the search term to a vector using the same embedding model. We perform a cosine similarity on the vector database to identify the appropriate context. Cosine similarity involves computing the dot product between the search vector and every knowledge vector in the database. OpenAI uses an absolute positional embedding, which means a cosine similarity of 1.0 means the context is an exact match to the search query. We then rank the contexts based on the dot product values (search term vs. knowledge vector). Once we have identified the top five contexts matching the search query, we feed those as a context to GPT API (we use a GPT generation model like \verb|text-davinci-003|) to create a customized answer. This context-specific GPT will guarantee deterministic answers and prevent hallucinations. Using this context-specific GPT, we get the correct answer ``The DIGGS XML tag for plastic limit is \textcolor{ForestGreen}{\textless diggs\_geo:waterContent\textgreater}within the parent tag \textcolor{ForestGreen}{\textless diggs\_geo:plasticLimitTrial\textgreater}” We can reuse the search engine to do more complex tasks, such as asking it to generate a DIGGS XML file (see~\cref{fig:diggs}). We can also finetune the semantic search engine using supervised learning trained on a list of questions and answers about DIGGS.

\subsection{Reasoning with a train of thought}
Can we make GPT do geotechnical analyses? \cite{wei2022chain} showed that instead of finetuning a separate language model for each new task, it is possible to just "prompt" the model with a handful of input-output examples to demonstrate the task. Let us now consider if we can provide examples to GPT to classify fine-grained soils based on liquid and plastic limits.
\\\\
\textit{\textcolor{RedOrange}{Prompt: Using soil classification of fine-grained soils as CL, CH, ML, or MH, classify the following soil with a Liquid Limit of 30, Plastic Limit of 10. }}
\\\\
\textcolor{blue}{\textit{Response}: In this case, PI = LL - PL = 30 - 10 = 20. Since the PI of the given soil is 20, it falls into the category of High Plasticity Soil. Therefore, the soil can be classified as either CH (High Plasticity Clay) or MH (High Plasticity Silt) based on visual and textural characteristics.}
\\GPT incorrectly classifies the soil as CH or MH when the liquid limit is less than 50\%. How can we build a reasoning engine that understands the underlying logic? We explore how reasoning abilities emerge naturally in large language models via a simple method called chain-of-thought prompting. Imagine solving a geotechnical engineering problem, like calculating the number of trucks needed to construct an embankment. We typically break down the problem into smaller steps and solve each before arriving at the final answer: "If the embankment requires 500 cubic meters of soil and each truck carries 25 cubic meters, we need 20 trucks... then, considering a 10\% compaction loss, we need an additional two trucks... so the total number of trucks required is 22." We show how to provide language models to classify soils with a similar thought process—a coherent series of intermediate reasoning steps leading to the answer. 
\\\\
\textit{\textcolor{RedOrange}{Prompt: Context: Fine-grained soil classes (``CL," ``CH," ``ML, ``MH")
\\Q: What is the classification of fine-grained soil with a LL = 60 and a PL = 50? 
\\A: LL boundary of 50 classifies if it is ``L" low plastic or ``H" high plastic. LL = 60 \textgreater 50, so ``H". PI = LL - PL = 60 - 50 = 10. Check if PI is above A-line: If PI \textgreater 0.73 * (LL - 20) = 10 \textgreater 0.73 * (60 - 20), if it is above the A-line, it is ``C" for clay, else ``M" for silt. It is not so ``M" silt. Soil is ``MH." 
\\Q: What is the classification of fine-grained soil with a LL = 30 and PL = 10?
}}
\\\\
\textcolor{blue}{\textit{Response A:}LL boundary of 50 classifies if it is ``L" low plastic or ``H" high plastic. LL = 30 \textless 50, so ``L". PI = liquid limit - plastic limit = 30 - 10 = 20. Check if PI is above A-line: If PI \textgreater 0.73 * (LL - 20) = 20 \textgreater 0.73 * (30 - 20). It is above the A-line, so ``C" clay. So the soil is ``CL."}
\\\\GPT now correctly classifies the soils as low-plastic clay, following a similar reasoning process. The train-of-thought prompting provides a reasoning ability to GPT and facilitates knowledge discovery. 

Let us consider another example, a task of assessing the safety of a gravity retaining wall. The wall has specific dimensions and is acting upon particular soil properties. We're to ensure the wall is safe against sliding, overturning, and bearing, applying a safety factor of 1.25. In train-of-thought reasoning, we will break down each step into separate prompts and provide sufficient context for GPT to solve the problem. For e.g., ``the first step involves evaluating the safety against sliding. This means we have to compute the ratio of the driving force (active soil pressure) to the resisting force (frictional resistance at the base of the wall). If the calculated factor of safety is above 1.25, it implies that the wall is safe against sliding. In this scenario, we're assuming passive pressure isn't considered, as it may be necessary to remove the earth in front of the wall for placing services, leaving the active pressure to be resisted only by the friction at the base of the wall." Similarly, we can provide detailed context for GPT to perform calculations to check the safety against overturning by asking it to calculate the total moment about the base center to find the eccentricity 'e' and state that If the eccentricity is within the middle third of the base width, it would imply that the wall is safe against overturning. Finally, prompt GPT to check for safety in bearing pressure by using Meyerhof's reduced base width approach. If the pressure exerted by the wall is less than the bearing capacity of the soil (considering the safety factor of 1.25), the wall would be considered safe against bearing failure. In order to guide the model in understanding these steps, prompt engineering must be carefully designed, with each step detailed out clearly. This would allow the GPT model to follow a logical sequence and provide a solution that aligns with the geotechnical reasoning required in these evaluations.

Although, train-of-thought process offers a reasoning engine with GPT, it is still limited by the formulas used by GPT, which may be hallucinated, leading to incorrect solutions. The next section explores how to build a robust engineering framework with GPT.

\subsection{GPT-enabled engineering workflows}

Let us now consider a complex geotechnical engineering workflow of calculating the maximum load supported by a foundation. Typically, most engineering workflows need several calculation stages. Each stage feeds into the next one, leading to the final result. Language models, like GPT, have limitations. They cannot solve these complex workflows using only prompt engineering. In such situations, GPT-like models might `hallucinate'. They might give answers that seem correct. However, these answers could mislead designers who don't have the necessary domain expertise. The following examples illustrate this issue.

\noindent\textit{\textcolor{RedOrange}{Prompt: Calculate the maximum load that can be supported on the clay layer at a depth of 5 m below the ground surface due to the leaning tower of Pisa with a diameter of 20 m.}}

\textcolor{blue}{\textit{Response}: The maximum load the clay layer can support at a depth of 5 m below the ground surface due to the leaning tower of Pisa can be calculated using plane strain bearing capacity theory.
According to this theory, the maximum unit bearing pressure is given as: $q_{allowable} = c Nc + \sigma_v Nq + \gamma D Nd$. $c$ = Cohesion of clay layer ($kN/m^2$), $\sigma_v = $ Overburden Pressure (kPa), and $\gamma D =$ Density of clay layer ($kN/m^3$). Assuming an average cohesion of \SI{1}{\kilo\newton\per\meter\squared}, an average overburden pressure of 100 kPa and an average density of \SI{18.92}{\kilo\newton\per\meter\cubed}, the bearing capacity factor can be determined using Mohr-Coulomb Failure criterion, resulting in the following $Nc = 15, Nq = 5, Nd = 3$. Therefore, $q_{allowable} = 15 \times 1 + 100 \times 5 + 18.92 \times 3 = 734.6$ kPa. Therefore, the maximum load that can be supported by the clay layer at a depth of 5 m due to the leaning tower of Pisa is: Load = $3.14 \times (20)^2 \times 734.6 = 6,037,295.4$ kN}\\

Despite multiple trials, GPT-3 incorrectly reports a maximum allowable load as 6,037 MN. Although GPT states that it needs to use the bearing capacity equation, the equation is incorrect missing factors such 0.5 in the last term and incorrectly assuming $\gamma D$ as Density of the clay layer. GPT also randomly predicts values of the bearing capacity factors resulting in an incorrect allowable bearing capacity of 734 kPa, which also has an arithmetic error (actual answer for that GPT calculation is 571.76 kPa). Finally, GPT incorrectly assumes the given diameter of the tower of 20 m to be the radius, and calculates the maximum load, again incorrect calculation, as 6,037 MN (correct answer for that calculation is 922 MN). The maximum allowable load is 98 MN using an undrained strength of 35 kPa in the clay layer from the soil report. Hence, GPT reports an answer that is 61 times more than the maximum allowable load. Using GPT without proper reasoning and engineering understanding would result in catastrophic engineering failures. 

How can we enable GPT to solve complex engineering workflows? To answer the question, we explore how humans solve problems. An extraordinary characteristic of human intelligence is the capability to flawlessly integrate task-oriented activities with verbal cognition (or inner speech) and maintaining a working memory. Consider the situation of assembling a piece of furniture. Amid any two distinct steps, we might utilize verbal reasoning to monitor our progress (``Now that I've tightened all the screws, I should attach the back panel"), to deal with unusual circumstances or modify the strategy according to the conditions (``I've run out of the right screws, so I'll have to use these other ones and buy replacements later"), and to identify when we need additional information (``What is the proper way to secure the drawer? I'll need to look that up online"). We may also take actions (read the assembly instructions, gather tools, verify parts) to facilitate the reasoning process and to respond to queries (``What part should I assemble next?"). This strong interplay between ``taking action" and ``reasoning" allows humans to rapidly acquire new skills and perform dependable decision-making or reasoning, even in novel situations or when faced with uncertain information.

\begin{figure}[!htbp]
    \centering
    \includegraphics[width=\textwidth]{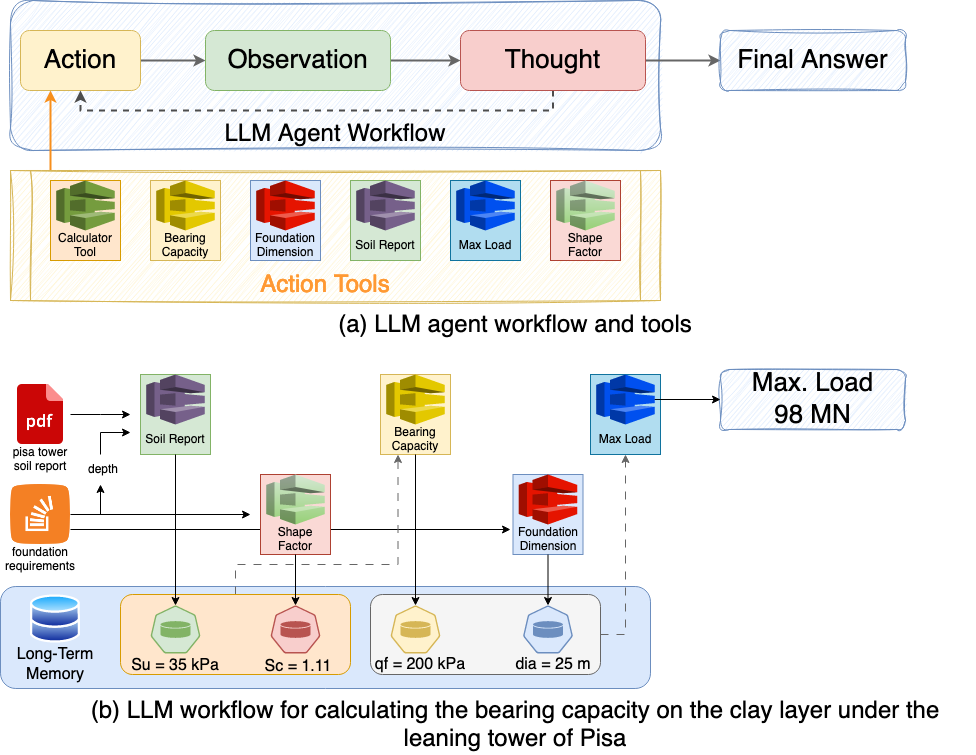}
    \caption{LLM Action-Observation-Thought workflow with short-term and long-term memory.}
    \label{fig:llm-workflow}
\end{figure}

The groundbreaking findings by~\textcite{yao2022react} on LLM ReAct suggest the potential of merging verbal reasoning with interactive decision-making within autonomous systems. We propose a novel strategy aimed at uniting the reasoning and linguistic capabilities of GPT (or inner speech) with a suite of high-level function libraries (action tools). This strategy addresses the challenges GPT models face, such as performing intricate geotechnical calculations like estimating bearing capacity. Instead of allowing GPT to perform these calculations unsupervised, potentially leading to inaccuracies, we introduce specific engineering tools. An example of such a tool is the `BearingCapacityTool', which inputs soil parameters and foundation dimensions to generate the allowable bearing pressure. By constraining the inputs, we let GPT know that in order to use the tool, it needs to provide two inputs and the output will be the bearing pressure. This allows GPT to string together a proper workflow, such as using the `SoilReportTool' to get soil parameters. However, this ReAct frameworks is constrained, and only supports short-term memory. This restriction means that knowledge from a previous step can only be transferred to the next immediate step. We propose a novel extension to the ReAct framework that incorporates a database to store pertinent information, acting as a long-term memory for the LLM workflows. All LLM Action tools can access this database, enabling them to utilize both short-term and long-term memories in their calculations. This unique integration of short and long-term memory, along with the LLM's linguistic prowess, creates a unified AI framework capable of solving complex engineering workflows. An overview of this framework, which includes an architecture of Action - Observation - Thought enabled by both Short and Long Term memory, is illustrated in~\cref{fig:llm-workflow}a. 

\Cref{fig:llm-workflow}b shows how GPT accesses the engineering tools using the Action-Observation-Thought framework to calculate the maximum load accurately for the same leaning tower problem. When provided with the task of calculating the load, GPT offers a reasoning workflow and identifies the relevant tools to use to solve the problem. GPT takes an action with the `SoilReportTool' to extract strength parameters, makes an observation, then follows it with more tools `BearingCapacityTool', `ShapeFactorTool' and finally the `MaxLoadTool' to extract the final answer. The GPT reasoning workflow is shown below.

\noindent\textit{\textcolor{RedOrange}{Prompt: Calculate the maximum load that can be supported on the clay layer at a depth of 5 m below the ground surface due to the leaning tower of Pisa with a diameter of 20 m using the soil report: pisa\_report.pdf.}}
\\
\textcolor{blue}{\textit{Response}: We need to use the soil report to extract the undrained strength parameters and then use that to calculate the bearing capacity and maximum load. We also need to save the foundation dimension.}\\

\noindent\textcolor{OliveGreen}{
\textbf{Action Tool}: SoilReport \\
\textbf{Action Input}: pisa\_report.pdf \\
\textbf{Action}: Extracting soil parameters from report:  pisa\_report.pdf \\
Interpolated strength of Clay from a depth of -5 to -17.5 m asl. Su = 35 kPa.
}\\

\noindent \textcolor{OrangeRed}{\textbf{Observation}: Su = 35 kPa}\\

\noindent \textcolor{blue}{\textbf{Thought}: Now that we have the soil report, we can use BearingCapacity to calculate the bearing capacity of the foundation.}\\

\noindent\textcolor{OliveGreen}{
\textbf{Action Tool}: BearingCapacity\\
\textbf{Action Input}: Su = 35 kPa and shape factor (Sc = 1.11) (from long-term database)\\
\textbf{Action}: Calculate bearing capacity of a circular foundation with soil parameters:  {`Su': 35.0}.
}\\

\noindent\textcolor{OrangeRed}{\textbf{Observation}: Bearing capacity = 199.689 kPa}\\

\noindent \textcolor{blue}{\textbf{Thought}: Now that we have the bearing capacity, we can use MaxLoad to calculate the maximum load.}

\noindent\textcolor{OliveGreen}{
\textbf{Action Tool}: MaxLoad \\
\textbf{Action Input}: Bearing capacity ($q_f = 199.689$ kPa and foundation dimension ($\phi = 20 $ m).\\
Transfer load to layer at 5 m depth using 2:1 stress transfer mechanism.}\\

\noindent\textcolor{OrangeRed}{\textbf{Observation}: Max. Load = 98022.02586093749 kN}\\

\noindent \textcolor{blue}{\textbf{Thought}: I now know the final answer.}\\

\noindent \textcolor{BrickRed}{\textbf{Final Answer}: The maximum load on the clay layer at 5 m depth for the leaning tower of Pisa with a diameter of 20 m is \textbf{98.022 MN}.\\}

\noindent Using the ReAct framework GPT accurately solves the maximum load using both short and long term memories. By developing tools to formulate workflows, GPT will revolutionize and accelerate engineering progress by serving as a unified input interface for complex tasks such as generating a mesh for finite element analysis, monitoring structural health, analyzing data or designing foundations, all requiring specialized skills. 

\subsection{The future of geotechnical engineering with LLMs}
LLMs, like GPT, offer numerous benefits to geotechnical engineering. Open-source LLMs, like LlaMA~\parencite{touvron2023llama} and Alpaca, require less computational power and democratize LLM access. However, it is crucial to develop context-specific models to maximize the potential of LLMs. This process involves accessing relevant data and constructing specialized interfaces tailored to the unique requirements of the field. While GPT can serve as an effective reasoning engine, it is essential to remember that context plays a significant role in its performance. Additionally, GPT has the potential to become a natural interface for completing complex tasks, such as data analysis and design. By integrating GPT into geotechnical engineering workflows, professionals can streamline their work and make informed decisions more efficiently, developing sustainable and resilient infrastructure systems of the future.
\section*{Acknowledgments}
I thank Prof. Ellen Rathje for her insightful comments and discussions in writing the article. 
\clearpage
\printbibliography
\end{document}